\documentclass[10pt,twocolumn,letterpaper]{article}

\usepackage{iccv,times,graphicx,tabulary,multirow,subfig,xspace,xcolor}
\usepackage{amsthm,amssymb,fixmath,mathtools,nicefrac,tabularx,booktabs}

\usepackage[pagebackref=true,breaklinks=true,letterpaper=true,colorlinks,bookmarks=false]{hyperref}
\newcommand{\tabincell}[2]{\begin{tabular}{@{}#1@{}}#2\end{tabular}}

\begin{document}
\title{MegDetV2}
\track{\emph{COCO Instance Segmentation}}

\author{Zeming Li, Yuchen Ma, Yukang Chen, Xiangyu Zhang, Jian Sun \\
Megvii Technology \\
{\tt\small \{lizeming, mayuchen, chenyukang, zhangxiangyu, sunjian\}@megvii.com}
}

\maketitle

\begin{abstract}

In this report, we present our object detection/instance segmentation system, MegDetV2, which works in a two-pass fashion, first to detect instances then to obtain segmentation. Our baseline detector is mainly built on a new designed RPN, called RPN++. On the COCO-2019 detection/instance-segmentation test-dev dataset, our system achieves 61.0/53.1 mAP, which surpassed our 2018 winning results by 5.0/4.2 respectively. We achieve the best results in COCO Challenge 2019 and 2020.
\end{abstract}

\section{Two-Pass Pipeline}
\begin{figure}
\includegraphics[width=8cm, height=8.5cm]{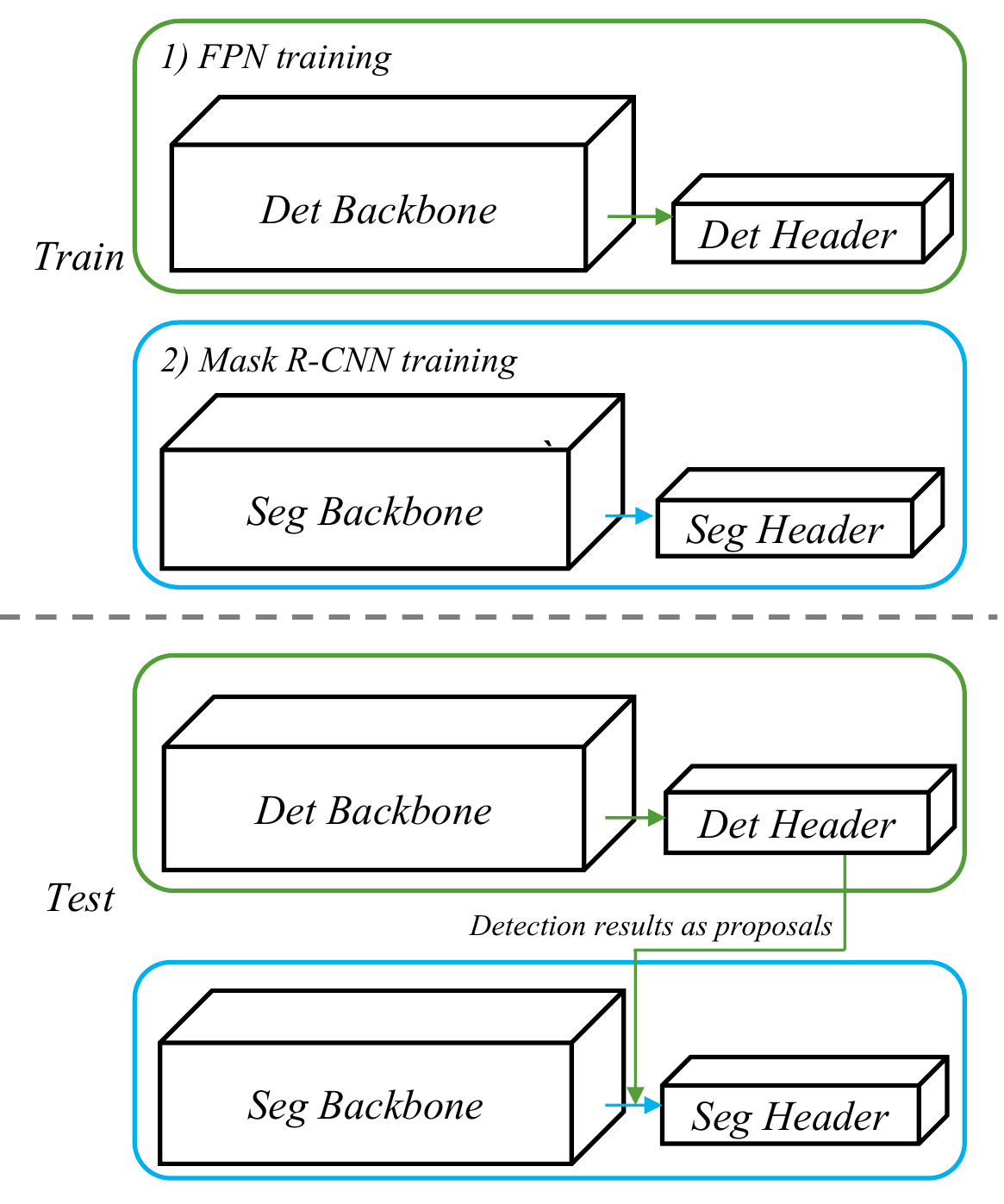}
\caption{Our two-pass pipeline for instance segmentation. We train FPN and Mask R-CNN for detection and instance-segmentation respectively. In test, we first generate the bounding boxes with FPN, then use them as proposals for Mask R-CNN. }
\vspace{-5pt}
\label{fig:two_pass}
\end{figure}

Following the COCO-2018-Megvii~\cite{coco2018_megvii}~(Fig.~\ref{fig:two_pass}) Two-Pass pipeline, we train a FPN detector~\cite{fpn} and a Mask-RCNN~\cite{mask_rcnn} separately. During testing, we feed the bounding boxes extracted from the FPN detector into the segmentation header of Mask-RCNN to generate final results. In this divide-and-conquer fashion, it is easy to leverage training data from different sources and manage very large-scale model training.

\section{Detection} \label{sec:detection}
\subsection{Training \&\& Inference} \label{sec:training}
We validate our system on the COCO-2017 datasets, which contain 135K images for training and 5K images for validation. We also report final results in the \emph{test-dev}. In the final submission, we leverage Objects365 detection datasets~\cite{objects365} for pre-training, which includes 365 object categories, and 600K training/30K validation images.



We use SGD with the momentum of 0.9 and 16 images per mini-batch for verification experiments. Learning rates are set to 2e-2 at the beginning of the training, reduced by the factor of 0.1 at 60K/20K/10K iterations respectively~(It is known as the Detectron~\cite{Detectron2018} `1x' settings).  

When we pre-train Objects365, we use 64 V100 GPUs~(2 images/GPU). Therefore 128 images are utilized per mini-batch. The learning late is linearly scaled to 1.6e-1, reduced by the factor of 0.1 at 40K/10K/5K.  Almost all experiments are conducted by Detectron `1x' schedule, because of the limitation of the time and computation. We also find larger models can be proficient with less time.


\begin{table*}[h]
\centering
\small
\caption{The results of combining different components.}
\setlength{\tabcolsep}{1.9mm}{
\begin{tabular}{l|ccccccc|cccccccc}
\toprule
FPN      & \tabincell{c}{deform \\conv}  & \tabincell{c}{deform \\pool } & \tabincell{c}{4conv \\gn} & \tabincell{c}{fpn \\ deform} & \tabincell{c}{high IoU \\ sample}  & \tabincell{c}{class aware \\ sample}  & \tabincell{c}{instance \\ seg}  & AP& AP$_{.5}$&AP$_{.75}$ & AP$_s$ & AP$_m$ & AP$_l$\\
\midrule
\checkmark    &   & &   &&&&& 36.3 & 58.8 & 39.1 & 22.4 & 39.5 & 45.6  \\ 
\checkmark & \checkmark &   & &    &&&& 38.9 & 61.7 & 41.8 & 24.4 & 41.8 & 50.8 \\ 
\checkmark & \checkmark & \checkmark  &&&&&& 			 40.1 & 62.6 & 43.2 & 24.8 & 43.2 & 52.5  \\ 
\checkmark & \checkmark & \checkmark & \checkmark &&&&& 40.8 & 63.2 & 44.0 & 25.8 & 43.9 & 54.2  \\ 
\checkmark & \checkmark & \checkmark & \checkmark &  \checkmark &&&& 41.1 & 63.4 & 44.4 &26.9 & 44.4 & 54.4  \\ 
\checkmark & \checkmark & \checkmark & \checkmark & \checkmark & \checkmark &&& 44.8 & 63.3 & 48.7 & 27.3 & 47.7 & 59.2  \\ 
\checkmark & \checkmark & \checkmark & \checkmark & \checkmark & \checkmark & \checkmark && 45.4 & 63.8 & 49.5 & 28.5 & 48.5 & 59.4 \\
\checkmark & \checkmark & \checkmark & \checkmark & \checkmark & \checkmark & \checkmark& \checkmark&  45.9 & 64.1 & 50.4 & 28.9 & 49.1 & 60.6  \\
\bottomrule
\end{tabular}
}
\label{tab:mask_rcnn_res}
\end{table*}

\subsection{Methods}
We will brief our methods in this subsection. All the experiments have done in Megvii internal deep learning platforms. We have re-implemented standard FPN~\cite{fpn}, Mask R-CNN~\cite{mask_rcnn} and Cascade R-CNN~\cite{cascade_rcnn}. Unless specified, we use COCO2017 data without Objects365 for training, and ResNet-50 is adopted as our basic feature extractor.

\subsubsection{RPN++} \label{sec:RPN++}
\textbf{High-IoU proposal sampling.} 
There are lots of high-quality proposals already exist in RPN, while we do not efficiently utilize them and filter them by NMS. Instead of using RPN scores to determine whether the proposals are good or bad. We propose to involve the IoU of these proposals over ground-truth boxes for selecting proposals which highly overlap with ground-truth boxes. The NMS IoU threshold is relaxed to keep more high-quality proposals. Benefited from the high-IoU sampler, we can even directly learn R-CNN with a higher critical IoU threshold of 0.7 and it significantly improves the FPN by 2.5~(38.8 vs 36.3 in Tab.~\ref{tab:less-head}). Combining cascade R-CNN is also helpful, we add additional R-CNN head with 0.7 IoU threshold, which further improves results to 40.7. Noticing it yields superior performances~(40.7 vs 40.2 in Tab.~\ref{tab:less-head}) with even reduced computations compared with plain cascade R-CNN.


\begin{table}[h]
\centering
\small
\caption{Comparison of FPN, Cascade R-CNN, and our proposed high-IoU proposal sampler.}  
\setlength{\tabcolsep}{1.2mm}{
\begin{tabular}{l|cccccc}
\toprule
Methods      &     AP & AP$_{.5}$ & AP$_{.75}$  & AP$_s$ & AP$_m$  & AP$_l$ \\
\midrule
FPN    &     36.3 & 58.8 & 39.1 & 22.4 & 39.5 & 45.6 \\
Cascade R-CNN & 40.2 & 58.9 & 44.1 & 23.0 & 43.6 & 52.5 \\
\hline
FPN + high IoU & 38.8 & 56.4 & 42.2 & 22.0 & 42.1  & 50.4 \\
Cascade + high IoU & 40.7 & 58.8 & 44.5 & 23.8 & 43.8 & 53.4 \\

\bottomrule
\end{tabular}
}
\label{tab:less-head}
\end{table}

\noindent \textbf{Class aware sampling.} We also involve class-wise balance sampling for RPN proposals.
Previously, the same IoU thresholds are applied over all classes; we change it dynamically in each class. Specifically, we will calculate a ratio $\alpha$ of how many proposals have higher IoU than 0.5 w.r.t ground-truth boxes for all classes.   Then we will sample proposals according to this $\alpha$ for each class. We also design another target matching rules for RPN anchors. Each ground-truth box will be forced to match a set of anchor boxes, e.g. top 35 anchors for each ground-truth box. These changes improve FPN by 1.3~(in Table~\ref{tab:Rpn++})

\begin{table}[h]
\centering
\small
\caption{Comparison of FPN with/without RPN++.}  
\setlength{\tabcolsep}{1.2mm}{
\begin{tabular}{l|cccccc}
\toprule
Methods      &     AP & AP$_{.5}$ & AP$_{.75}$  & AP$_s$ & AP$_m$  & AP$_l$ \\
\midrule
FPN    &     36.3 & 58.8 & 39.1 & 22.4 & 39.5 & 45.6 \\
+class aware sample & 37.6 & 59.9 & 40.4 & 23.2 & 41.2 & 48.3 \\
\bottomrule
\end{tabular}
}
\label{tab:Rpn++}
\end{table}

\subsubsection{Strong-Baseline for Object Detection}
We further build a stronger baseline to achieve more competitive performance. The effectiveness of high-IoU sample and class aware sample is also verified on our strong baseline. We carefully study the recent developments in object detection and finally we adopt the technique of \emph{``Deformable Network v2 with pooling''}~\cite{deformablev2}, \emph{``Stacking 4 Convolutions for location branch''}. 

\paragraph{Deformable Convolution and Pooling.} 
Following the Deformable ConvNets v2, we add the deformable-conv into each bottleneck $3\times3$ convolution within ResNet stage-\{3, 4, 5\}. It yields 2.6~(38.9 vs 36.3) improvements. We further replace the RoI-Align operation with the deformable RoI-Align like Deformable ConvNets~\cite{deformable} does. The results become 1.1~(40.0 vs 38.9) higher.

\paragraph{Stacking 4 Convolutions for location branch.} 
Following the location-sensitive header proposed in COCO2018~\cite{coco2018_megvii}, instead of using 2 fully-connected layers to predict bounding box regression, we apply 4 stacked convolutions to better exploit \emph{spatial information} for location task. It effectively improves the results by 0.7~(40.8 vs 40.1) points. 

\paragraph{Feature Pyramid with deformable convolution.} 
The plain deformable convnet adopts deform-conv for backbone feature extraction, followed by feature pyramid fusion to combine the low-level details and high-level semantic features. There is no ``deformation'' after the backbone feature fusion. we make a simple design choice that changing the lateral $3 \times 3$ convolution within FPN to deformable $3 \times 3$ convolution. This leads 0.3~(41.1 vs 40.8) improvements with a simple modification.

\paragraph{High IoU sample.}
As mentioned in section~\ref{sec:RPN++}, High IoU sampler is quite efficient and effective in improving the FPN baseline. We further conduct the experiments to verify the sampler when combining FPN with other improvements like Deformable Convolution and 4-conv head for location branch. High IoU sampler with cascade r-cnn yields 3.8 improvements over a much stronger baseline, which shows the generalization capability for the detection task.

\paragraph{Class aware sample.}
Further adding class aware sampler proposed in \ref{sec:RPN++} improves the results by 0.6, it shows class aware sample can also benefit the stronger baselines.

\paragraph{Adding instance segmentation branch in R-CNN.}
Besides the detection, COCO dataset provides additional instance segmentation annotations, which enable more supervision to learn the general feature representations. Following the Mask R-CNN, we attach the instance segmentation branch in R-CNN and it gives 0.5~(45.9 vs 45.4) improvements over object detection.

\subsection{COCO 2019 Detection Road-Map.} \label{sec:det_road_map}
Finally, to gain better results for COCO challenges, we adopt another 3 powerful backbones to extract image features. There are SENet-154~\cite{senet}, Shuffle V2~\cite{shufflev2} and ResNext~\cite{resnext} with similar flops, and their performances are close to each other. We mainly build our ablations based on Shuffle V2.

The road map of the COCO2019 challenge is reported in Table~\ref{tab:coco2019_roadmap}.

\begin{enumerate}
\item Using stronger ShuffleNet V2 backbone achieves 5.2 improvements. Noticing, to speed up our experiments, we use 64 GPUs~(2 images/gpu ) when training larger backbones.
\item Adding SoftNMS leads to 0.7 improvements.
\item Furthering, we utilize the large batch~(128 images) synchronized Batch-Normalization~\cite{megdet}, which brings 0.9 gains. 
\item The multi-scale training strategy is involved in boosting our performance, specifically we determine the short-size of images by uniform sampling from the range of 400 to 1400, and the max size is limited to 1400. Compared with single-scale training, multi-scale training yields 0.9 improvements.
\item The larger detection datasets Objects365 is adopted as the pre-train datasets. We train Objects365 following \ref{sec:training} which utilizes 64 GPUs with synchronized batch-normalization. During COCO fine-tuning, we use a smaller learning rate, which is 0.15 $\times$ 1.6e-1 when the beginning of the training, reduced by factor 0.2 at 5.5K/7.5K/8K iterations. Pre-train Objects365 gives us 2.3 improvements. We only pre-train objects365 with Detectron `1x' setting, longer pre-train has not been tested.
\item We choose a larger size range of 600-1600 for fine-tuning COCO, the max size for longer edge is limited to 1867. Compared to plain sizes of 400-1400 for multi-scale training, it yields 0.9 gains. We have not tried a bigger size because it increases much training time.
\item We further extend the time for multi-scale training, involving larger RoI as context, as well as normalizing the classification score of R-CNN. These `tricks' improve the results by 1 point.
\item We use the larger sizes of \{600, 800, 1000, 1200, 1400\} for multi-scale testing. Horizontal flipping is applied for each scale. It further raises the results by 1.6.
\item We finally ensemble three different models, the results reach 60.7 in COCO2017 validation datasets, 61.0 in COCO2019 \emph{test-dev} datasets.
\end{enumerate}

\begin{table}[h]
\centering
\small
\caption{Detection road maps of COCO 2019 challenge .}  
\setlength{\tabcolsep}{1mm}{
\begin{tabular}{l|cccccc}
\toprule
Methods      &     AP & AP$_{.5}$ & AP$_{.75}$  & AP$_s$ & AP$_m$  & AP$_l$ \\
\midrule
Strong baseline & 45.9 & 64.1 & 50.4 & 28.9 & 49.1 & 60.6 \\
+ 1. shuffle v2    & 51.1 & 69.9 & 55.6 & 33.2 & 55.1 & 66.7 \\
+ 2. softnms~\cite{softnms}       & 51.8 & 69.5 & 57.2 & 33.6 & 55.7 & 67.4 \\
+ 3. syncbn~\cite{megdet}        & 52.7 & 70.7 & 58.2 & 34.6 & 56.8 & 57.5 \\
+ 4. ms train      & 53.6 & 71.6 & 59.2 & 36.2 & 57.4 & 68.7 \\
+ 5. Objects365~\cite{objects365}    & 55.9 & 74.1 & 61.7 & 39.8 & 60.6 & 69.9 \\
+ 6. larger size   & 56.8 & 74.8 & 62.7 & 40.2 & 60.8 & 69.8 \\
+ 7. tricks        & 57.8 & 75.9 & 64.0 & 43.2 & 62.2 & 71.2 \\
+ 8. ms testing    & 59.4 & 77.5 & 66.0 & 45.3 & 63.8 & 72.9 \\
+ 9. ensemble      & 60.7 & 78.9 & 67.4 & 46.1 & 64.8 & 73.6 \\
\hline
ensemble test-dev & 61.0 & 79.1 & 67.9 & 43.9 & 64.2 & 73.9 \\
\bottomrule
\end{tabular}
}
\label{tab:coco2019_roadmap}
\end{table}

\section{Segmentation} \label{sec:instance_segmentation}

\subsection{Methods}
We adopt the location-sensitive header~\cite{coco2018_megvii} as our baseline. Unless specified, experiments are conducted with ResNet-50 backbone.

\paragraph{Mask Rescore.} 
Following the Mask Scoring~\cite{mask-rescore}, we add an additional Mask-IoU head to learn the quality of the predicted instance masks. Considering the speed/accuracy trade-off, we only use two $3 \times 3$ convolutional layers for mask-IoU predictions.

\paragraph{Stuff supervision.} 
The scene context provides useful clues for in instance segmentation. We utilize another fully convolutional branch to predict the stuff segmentation, which involves additional supervision to learn better feature representations.

\paragraph{Efficient Mask Cascade.} 
To accurately predict the instance segmentation, we involve additional mask head to learn the residual loss from the previous mask prediction. Noticing, we only refine the mask~(without bounding box refinement).


\subsection{COCO 2019 Instance Segmentation Road Map.}
Our final results are conducted by two powerful backbones including Shuffle V2~\cite{shufflev2} and ResNext~\cite{resnext}~(same as detection). In our ablations, most experiments are based on Shuffle V2. The detail road map for instance segmentation are listed as follows:

\begin{enumerate}
\item Mask rescore head achieves 0.6 improvements. 
\item Stuff supervision leads to 0.5 improvements. 
\item Efficient Mask Cascade yields 0.6 gains. 
\item We further adopt the ShuffleNet V2 as the backbone and utilize multi-scale during the training. The training sizes are randomly sampled from \{600, 800, 1000, 1200\}. Meanwhile, horizontal flipping is also applied for each scale. The results are improved by 7.4.
\item Using the best detection results also benefits for instance segmentation. By feeding the bounding boxes with 60.7 AP into the segmentation branch, the results are boosted by 5.7.

\item Multi-scale and horizontal flip testing is similar to detection. They further raise the results by 0.5.
\item We finally ensemble two different models, the results are 52.7 in COCO2017 validation datasets, 53.1 in COCO2019 \emph{test-dev} datasets.

\end{enumerate}

\begin{table}[h]
\centering
\small
\caption{Instance segmentation road map of COCO 2019 challenge.}  
\setlength{\tabcolsep}{1mm}{
\begin{tabular}{l|cccccc}
\toprule
Methods      &     AP & AP$_{.5}$ & AP$_{.75}$  & AP$_s$ & AP$_m$  & AP$_l$ \\
\midrule
COCO2018 baseline~\cite{coco2018_megvii}              & 36.8 & 57.9 & 39.8 & 20.3 & 39.4 & 49.6 \\
+ 1.mask rescore               & 37.4 & 57.8 & 40.6 & 21.3 & 40.2 & 50.7 \\
+ 2.stuff supervision          & 37.9 & 58.1 & 41.1 & 21.7 & 40.3 & 51.2 \\
+ 3.efficient mask cascade       & 38.5 & 58.2 & 43.1 & 23.6 & 42.7 & 52.3 \\
+ 4.shuffle v2 + ms-train      & 45.9 & 65.4 & 50.2 & 29.3 & 47.9 & 58.3\\
+ 5. two-pass pipeline & 51.6 & 76.0 & 56.5 & 37.0 & 55.6 & 66.3 \\
+ 6.ms testing                 & 52.1 & 76.4 & 57.1 & 37.8 & 56.1 & 66.5 \\
+ 7.ensemble                   & 52.7 & 76.5 & 58.2 & 36.1 & 56.1 & 67.5 \\
\hline
ensemble test-dev  & 53.1 & 76.8 & 58.6 & 36.6 & 56.5 & 67.7 \\
\bottomrule
\end{tabular}
}
\label{tab:coco2019_seg_roadmap}
\end{table}



{\small \bibliographystyle{ieee_fullname} \bibliography{egbib}}

\end{document}